\documentclass[conference]{IEEEtran}

\usepackage{cite}
\usepackage{amsmath,amssymb,amsfonts}
\usepackage{algorithmic}
\usepackage{graphicx}
\usepackage{textcomp}
\usepackage{xcolor}
\usepackage{booktabs}
\usepackage{tabularx}
\usepackage{url}
\usepackage{comment}
\usepackage[hidelinks]{hyperref}

\def\BibTeX{{\rm B\kern-.05em{\sc i\kern-.025em b}\kern-.08em
    T\kern-.1667em\lower.7ex\hbox{E}\kern-.125emX}}

\begin{document}

\title{TripleGrader: A Web-Based Platform Supporting Collaborative Triple Annotation with Quality Grading and Inter-Rater Agreement Assessment}

\author{
\IEEEauthorblockN{Rajan Lamichhane}
\IEEEauthorblockA{\textit{Department of Computer Science} \\
\textit{Charleston Southern University}\\
Charleston, SC, USA\\
rlamichhane@student.csuniv.edu}
\and
\IEEEauthorblockN{Songhui Yue}
\IEEEauthorblockA{\textit{Department of Computer Science} \\
\textit{Charleston Southern University}\\
Charleston, SC, USA\\
syue@csuniv.edu}
}

\maketitle
{\let\thefootnote\relax\footnotetext{This work has been submitted to the IEEE for possible publication. Copyright may be transferred without notice, after which this version may no longer be accessible.}}

\begin{abstract}
Knowledge graphs need validated relational triples, and gold benchmarks of such triples are often built by several annotators. Most text annotation platforms focus on span tagging and compute agreement on categorical labels. They are not designed for ordinal quality grading of triples that have already been extracted. We present TripleGrader, an open-source web platform for grading subject-predicate-object triples on a 0 to 10 ordinal scale. In TripleGrader, each triple is scored independently, and peer scores stay hidden until the grader has rated it. Graders then discuss entries in threaded comments and propose revisions, which the session owner approves and publishes as new versions. For agreement, the platform computes quadratic-weighted Cohen's kappa for two raters and a quadratic-weighted Fleiss-type kappa for three or more. Graded sessions and the final benchmark are exported in common formats. We evaluated the platform in three ways. We ran 72 automated tests, including 18 workflow scenarios through the REST API. We also injected 10 error types to check whether prescribed corrections could be applied while preserving unaffected triples. On two small corpora with existing gold benchmarks, we then used the platform to curate the triples extracted by four large language models into each benchmark. Only 10\% to 32\% of the candidate triples could be kept unchanged, and 53\% to 71\% needed corrections, partly to match the benchmarks' naming conventions. Per model, 11 to 29 missing triples had to be added. On these corpora, the extracted triples therefore needed substantial human review. The source code is available at \url{https://github.com/razz0n/triple-grader-tool}.
\end{abstract}


\section{Introduction}
\label{sec:introduction}

Relational triples, or subject-predicate-object (SPO) triples, are a common way to represent structured knowledge in knowledge graphs~\cite{hogan2021knowledge} and ontologies~\cite{gruber1993translation,noy2001ontology,yue2024csm}. Natural language processing (NLP) systems can extract candidate triples from text automatically~\cite{angeli2015leveraging}, although their output often contains wrong or unsupported relations~\cite{mihindukulasooriya2023text2kgbench}. For this reason, validated knowledge graphs still need human review~\cite{hogan2021knowledge}.

Text annotation platforms such as brat~\cite{stenetorp2012brat}, Label Studio~\cite{tkachenko2020labelstudio}, and INCEpTION~\cite{klie2018inception} are organized around marking spans and relations in raw text. Scoring pre-extracted triples on an ordinal scale, and then computing weighted agreement on those scores, falls outside their main workflow (Section~\ref{sec:related_work}). A team that grades triples with these tools has to export the scores and compute agreement with separate scripts. The process is then hard to repeat, and the resulting datasets are hard to reuse.

This paper asks whether a dedicated platform can support the whole cycle of triple grading, from independent scores to a published benchmark, and how much human curation triples extracted by large language models need before they match a gold benchmark. We present \emph{TripleGrader}, a web platform for grading pre-extracted relational triples. A session owner uploads candidate triples, which independent graders score on a 0 to 10 scale and discuss. The owner then approves revisions and publishes versions. The paper makes three contributions: (i) the platform, with blind first-round grading, re-grading of revised triples, and quadratic-weighted agreement for two or more raters; (ii) an evaluation of its workflows, including recovery from injected errors; and (iii) a curation study on two corpora in which the output of four language models was edited into an existing gold benchmark.

Section~\ref{sec:motivation} motivates the work and Section~\ref{sec:related_work} reviews related tools and agreement measures. Sections~\ref{sec:architecture} and~\ref{sec:capabilities} describe the architecture and the features of the platform, with a worked example and its field use. Section~\ref{sec:evaluation} reports the evaluation, Section~\ref{sec:discussion} discusses applications and limitations, and Section~\ref{sec:conclusion} concludes.

\section{Motivation}
\label{sec:motivation}

\subsection{Why Triple Annotation Is Needed}
Knowledge graphs encode facts as SPO triples and are used for search and question answering~\cite{hogan2021knowledge}. An error in a triple can therefore affect any application that reads the graph.

Candidate triples can be generated at scale by information extraction pipelines~\cite{angeli2015leveraging,manning2014stanford} and large language models (LLMs)~\cite{yue2026llm,he2024zero}. Their output is noisy. Some triples keep unresolved pronouns or state relations that the source text does not contain~\cite{mihindukulasooriya2023text2kgbench}.

Human review is therefore the usual way to validate extracted triples~\cite{kontokostas2013triplecheckmate,acosta2013crowdsourcing}. Annotators read the source sentence and judge whether each candidate represents a relationship stated in the text accurately and completely.

\subsection{Current State of Triple Annotation}
The tooling available for this kind of review is limited (Section~\ref{sec:related_work}). INCEpTION computes agreement for categorical annotations. Triple review can instead use an ordinal grade (for example, 0 to 10) to record partial correctness. Such grades call for weighted agreement coefficients. When we reviewed LLM-extracted triples in INCEpTION and Label Studio, we had to export the scores and compute quadratic-weighted kappa with offline scripts. Refining a triple also needs item-level discussion and repeated revision. Tagging platforms do not tie these steps to the scores.

\begin{figure}[htbp]
    \centering
    \includegraphics[width=0.9\columnwidth, keepaspectratio]{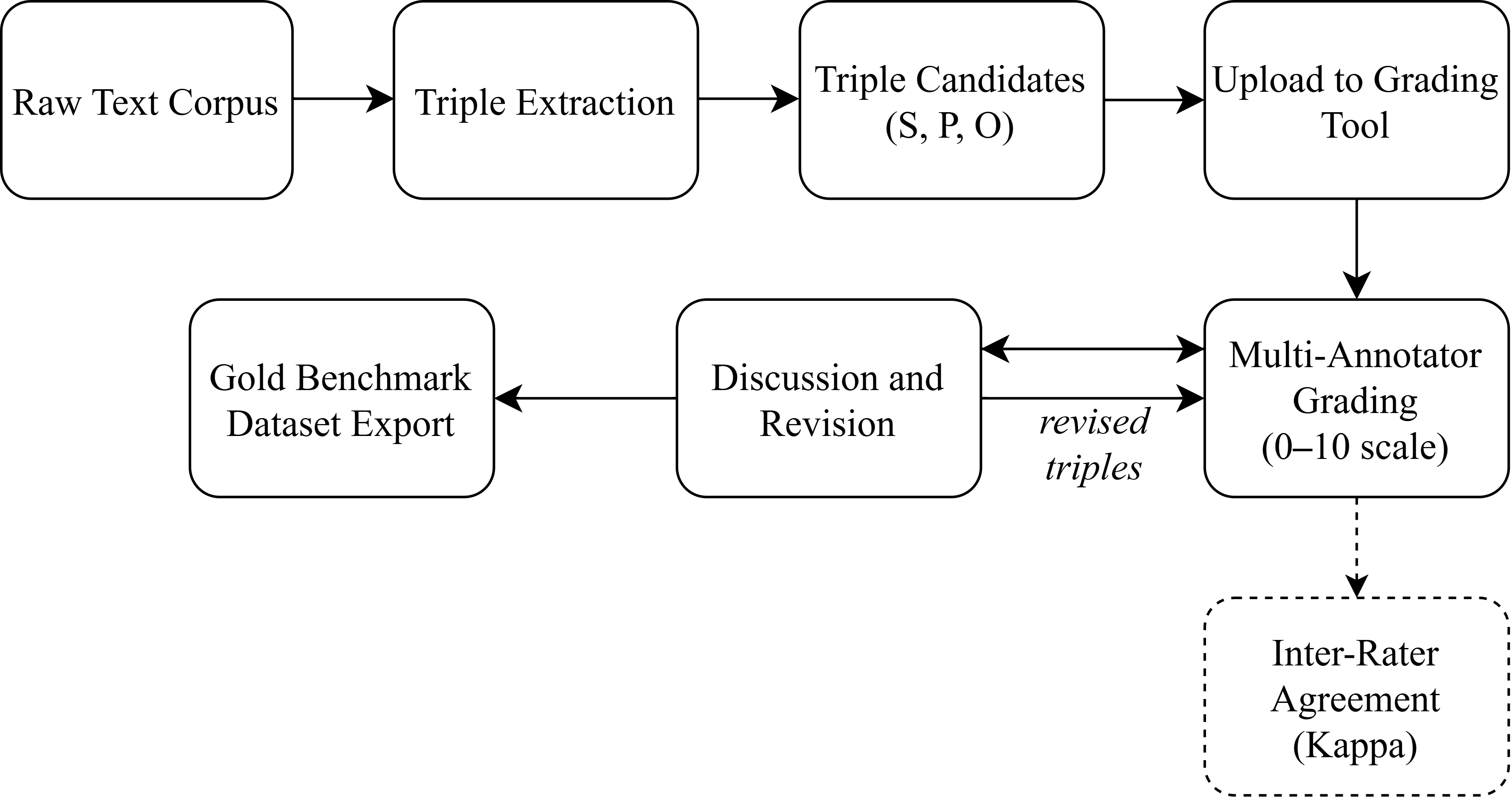}
    \caption{Annotation and benchmark creation pipeline.}
    \label{fig:pipeline}
\end{figure}

\subsection{The Benchmarking Gap}
Benchmarks such as Text2KGBench~\cite{mihindukulasooriya2023text2kgbench} exist for triple extraction, but building a new one for a specific domain is still largely ad hoc. We designed TripleGrader to make that process repeatable. In a session, one grading scale is applied to all triples, and agreement and version history are stored with it. The result is exported as a structured benchmark. Fig.~\ref{fig:pipeline} shows the pipeline from raw text to gold benchmark export.

\section{Related Work}
\label{sec:related_work}

Methodologies for building domain ontologies describe an iterative process in which the model is drafted, evaluated, and revised~\cite{noy2001ontology}. Ontology engineering environments such as Prot\'eg\'e support Web Ontology Language (OWL) authoring and formal reasoning~\cite{musen2015protege}. Their focus is logical modeling. The task we address is the scoring of extracted candidates by several annotators.

\subsection{Text Annotation Platforms}
A recent survey by Colucci Cante et al.~\cite{colucci2024text} compares text annotation tools. brat~\cite{stenetorp2012brat} renders entity spans, relations, and events over the text and can show two annotation sets side by side, but it computes no agreement statistic. Label Studio~\cite{tkachenko2020labelstudio} labels text, images, audio, and video. Text relations are links between labeled regions, and rating controls can record a score per task. Label Studio offers agreement metrics only in its commercial edition. INCEpTION~\cite{klie2018inception} supports entity linking, imported pre-annotations, recommenders, and curation. For categorical annotations, it computes Cohen's and Fleiss' kappa, Krippendorff's alpha~\cite{artstein2008inter}, and Gwet's AC1 and AC2. Of these tools, it is the closest to our purpose, although its workflow centers on spans and relations in text. GATE Teamware~\cite{bontcheva2013gate} assigns workflow roles and distributes documents across annotators, and its curation step computes agreement and lets a curator reconcile differences. Table~\ref{tab:tool_comparison} compares TripleGrader with brat, Label Studio, INCEpTION, and Potato, which is discussed below.

\begin{table*}[t]
\caption{Built-In Features Relevant to Grading Pre-Extracted Triples}
\label{tab:tool_comparison}
\centering
\footnotesize
\setlength{\tabcolsep}{4pt}
\begin{tabularx}{\textwidth}{@{}>{\raggedright\arraybackslash}p{2.6cm}*{5}{>{\raggedright\arraybackslash}X}@{}}
\toprule
\textbf{Feature} & \textbf{TripleGrader} & \textbf{brat}~\cite{stenetorp2012brat} & \textbf{Label Studio}~\cite{tkachenko2020labelstudio} & \textbf{INCEpTION}~\cite{klie2018inception} & \textbf{Potato}~\cite{pei2022potato} \\
\midrule
Primary task & Grading SPO triples & Span/relation tagging & Multi-modal labeling & Span/relation annotation & Configurable item annotation \\
Input unit & Pre-extracted triples & Raw text & Text, image, audio, video & Text with optional pre-annotations & Text, image, audio, video items \\
Ordinal score per item & 0--10 grade & No & Rating control & Numeric features & Likert, slider \\
Built-in agreement & Cohen/Fleiss-type $\kappa$ & No & Commercial edition & Cohen, Fleiss, Krippendorff, Gwet & Cohen, Fleiss, Krippendorff \\
Weighted agreement for ordinal scores & Quadratic & No & No & No & Linear, quadratic \\
Grades dropped when a triple is revised & On version publish & n/a & No & No & No \\
Export formats & CSV, JSON, XLSX, DOCX, PDF & Standoff (.ann) & JSON, CSV, others & UIMA CAS XMI, TSV, others & JSON, CSV, Parquet, CoNLL, others \\
\bottomrule
\multicolumn{6}{@{}p{\textwidth}@{}}{\scriptsize Open-source editions, from each tool's documentation in October 2026; ``Commercial edition'' marks a feature offered only in Label Studio's paid edition; n/a: the tool records no per-item score.}
\end{tabularx}
\end{table*}

\subsection{From Span Tagging to Ordinal Triple Grading}
As Table~\ref{tab:tool_comparison} shows, most of these platforms treat annotation as marking a span and choosing a category label. Reviewing the output of extraction systems or LLMs is different work. Here the extraction already exists, and the reviewer judges whether each triple is faithful to the sentence and complete.

These judgments are ordinal. Nominal kappa still counts a difference between grades 9 and 8 as a full disagreement, the same as a difference between 9 and 2. For graded judgments of this kind, Artstein and Poesio~\cite{artstein2008inter} describe weighted coefficients that give partial credit to near agreement. Some general-purpose tools now offer such measures for rating scales. The current release of Potato~\cite{pei2022potato} (version 2.13) reports linear and quadratic weighted kappa and ordinal Krippendorff's alpha for Likert items, and version 1 of Argilla's Python client~\cite{vilasuero2023argilla} computes alpha for rating questions. Linked Data quality assessment also reviews triples one at a time. In TripleCheckMate~\cite{kontokostas2013triplecheckmate} and in the crowdsourcing workflow of Acosta et al.~\cite{acosta2013crowdsourcing}, reviewers assign categorical error types to DBpedia triples. To our knowledge, none of these tools treats a triple as an item whose wording graders can revise with the owner's approval and whose grades are dropped when that wording changes. In TripleGrader, we combined ordinal grading and quadratic-weighted Cohen's~\cite{cohen1968weighted,fleiss1973equivalence} and Fleiss-type kappa~\cite{fleiss1971measuring} with this revision and re-grading workflow.

\section{System Architecture}
\label{sec:architecture}

\begin{figure}[htbp]
    \centering
    \includegraphics[width=\columnwidth]{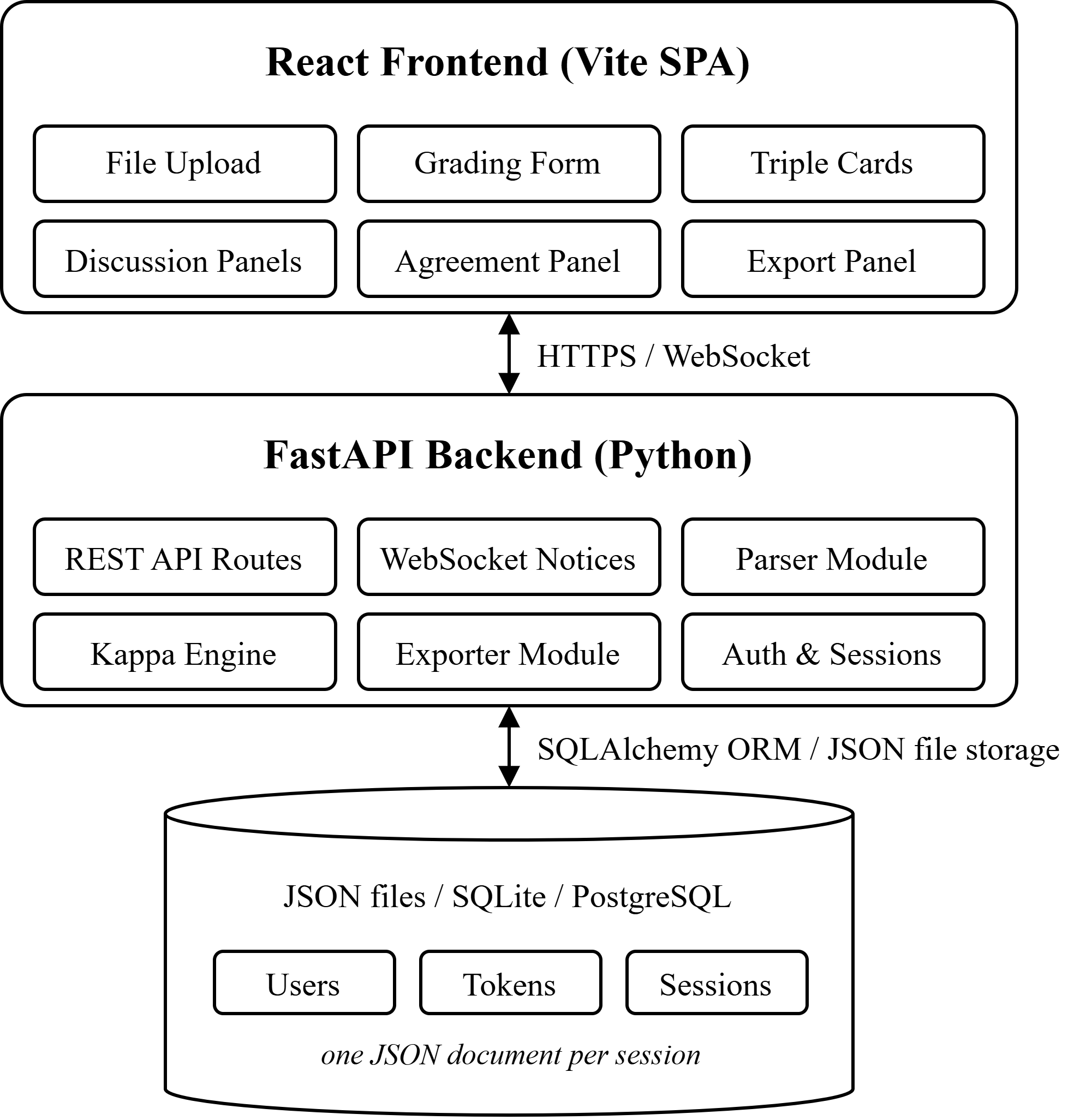}
    \caption{Three-tier system architecture with frontend, backend, and database layers.}
    \label{fig:architecture}
\end{figure}

We built TripleGrader in three tiers (a browser frontend, a backend web service, and a database). Fig.~\ref{fig:architecture} shows how they communicate.

\subsection{Frontend Layer}
The frontend is a single-page application written in React and built with Vite. Its components cover file upload, the grading form with its triple cards, discussion panels, the agreement panel, and export.
\subsection{Backend Application Layer}
The backend is written in Python with FastAPI. It exposes REST routes for authentication, file parsing, session management, agreement computation, and report generation. We hash passwords with PBKDF2-HMAC-SHA256 and a per-user salt. After login, a client receives a random bearer token, which the server stores and expires after six hours of inactivity. The routes that read or change a session, including the WebSocket channel, require this token. The server identifies the acting user from the token alone. When a grader saves the session, the server keeps only the changes that grader may make (their own scores, comments, suggestions, votes, removal requests, and proposed triples) and takes everything else from the stored copy. Connected clients get a WebSocket notification when a session is saved or a version changes. Other updates, such as new comments, reach them through periodic polling.

\subsection{Database Storage Layer}
In the deployed system, users, tokens, and sessions are stored in PostgreSQL (hosted on Supabase) through SQLAlchemy. SQLite can take the same role on a single machine. A default local installation keeps these records as JSON files on disk. We store each session as one JSON document. It holds the sentences, candidate triples, per-user ratings, comment threads, suggestions, and published versions of the session. The full state of a session is therefore saved and exported as a unit.

\section{System Usage and Capabilities}
\label{sec:capabilities}

A grading session can be created in two ways. In \textbf{file upload mode}, the owner submits a CSV, Excel, or JSON file. For spreadsheets, the parser reads subject, predicate, and object from the first three columns and groups the triples under their source sentences. JSON files may contain sentence arrays with their triple lists directly. In \textbf{manual builder mode}, the owner types source sentences and their triples in the browser.

An active session provides the following features:
\begin{enumerate}
    \item \textbf{Triple grading}: graders score each triple from 0 to 10. Each grade is stored under the grader's user ID. Every grader therefore keeps an independent record. The session owner does not grade.
    \item \textbf{Sentence-level ratings}: each sentence is rated by graders from 0 to 10 for \emph{completeness} (how fully its triples capture the meaning of the sentence) and \emph{external relation} (whether the predicates match the schema properties and domain guidelines).
    \item \textbf{Blind first round}: until a grader has rated an item, the server withholds the other graders' scores and averages for it.
    \item \textbf{Discussion and suggestions}: users comment on sentences and triples in threaded discussions. Graders propose revised or new triples and vote on other graders' proposals. An approved proposal replaces the triple's wording, and the owner can undo the approval. Proposals left unapproved are cleared when the next version is published.
    \item \textbf{Versions and re-grading}: each grade records the wording of the triple it was given for, and each sentence-level rating records the set of triples it was given for. When the owner publishes a named version, the server drops grades whose triple wording has since changed and flags the triple with a description of the change. Graders then score the new wording. Sentence-level ratings are dropped in the same way when the triples of their sentence have changed. The published version becomes the active version for the next round. Earlier versions stay available read-only unless the owner deletes them.
    \item \textbf{Agreement}: the platform reports quadratic-weighted Cohen's kappa when two graders rated the items and a quadratic-weighted Fleiss-type kappa when three or more did (Section~\ref{subsec:irr}).
    \item \textbf{Export}: sessions export to CSV, Excel, Word, PDF, and JSON, and a separate export contains only the benchmark triples (Section~\ref{subsec:benchmark_construction}).
\end{enumerate}

\subsection{Inter-Rater Agreement Formulation}
\label{subsec:irr}
Grades are integers from 0 to $N$, with $N = 10$. Quadratic weighting penalizes large differences between grades more than small ones. The disagreement weight for grades $i$ and $j$ is
\begin{equation}
w_{i,j} = \frac{(i - j)^2}{N^2}.
\label{eq:quadratic_weight}
\end{equation}

For two raters $R_1$ and $R_2$, the quadratic-weighted Cohen's kappa~\cite{cohen1968weighted} is
\begin{equation}
\kappa_w = 1 - \frac{\sum_{i=0}^{N} \sum_{j=0}^{N} w_{i,j} O_{i,j}}{\sum_{i=0}^{N} \sum_{j=0}^{N} w_{i,j} E_{i,j}},
\label{eq:cohen_kappa}
\end{equation}
where $O_{i,j}$ is the observed proportion of items that $R_1$ graded $i$ and $R_2$ graded $j$, and $E_{i,j} = p^{(1)}_i p^{(2)}_j$ is the proportion expected by chance from each rater's own grade distribution over the co-rated items.

When three or more raters grade a session, each item $t$ rated by $m_t \ge 2$ raters contributes its mean pairwise disagreement. Chance disagreement is computed from the pooled distribution $p$ of all grades given on these items:
\begin{equation}
\begin{aligned}
\kappa_{F} &= 1 - \frac{\bar{D}_o}{D_e}, \qquad D_e = \sum_{i,j} w_{i,j}\, p_i\, p_j,\\
\bar{D}_o &= \frac{1}{n}\sum_{t=1}^{n} \frac{\sum_{a<b} w_{x_{ta},x_{tb}}}{\binom{m_t}{2}}.
\end{aligned}
\label{eq:fleiss_kappa}
\end{equation}
Here $n$ is the number of items rated by at least two raters, $x_{ta}$ is the grade that rater $a$ gave item $t$, and the inner sum runs over all pairs of raters who graded item $t$. This coefficient generalizes Fleiss' kappa~\cite{fleiss1971measuring}, which assumes unweighted categories and the same number of raters per item, to quadratic weights and to items with different numbers of raters, in the form of the weighted multi-coder coefficients described by Artstein and Poesio~\cite{artstein2008inter}. It is close to Krippendorff's alpha with the interval metric~\cite{artstein2008inter}, which weights items by their number of ratings and adds a small-sample correction. The platform reports kappa as not available in two cases. If all grades are identical ($D_e = 0$), the coefficient is undefined. If fewer than two items are co-rated, the value would depend only on the number of raters. The workspace shows kappa for triple grades and for each sentence-level rating, and the PDF report includes the same values.

\subsection{Worked Example}
\begin{figure}[htbp]
    \centering
    \includegraphics[width=\columnwidth]{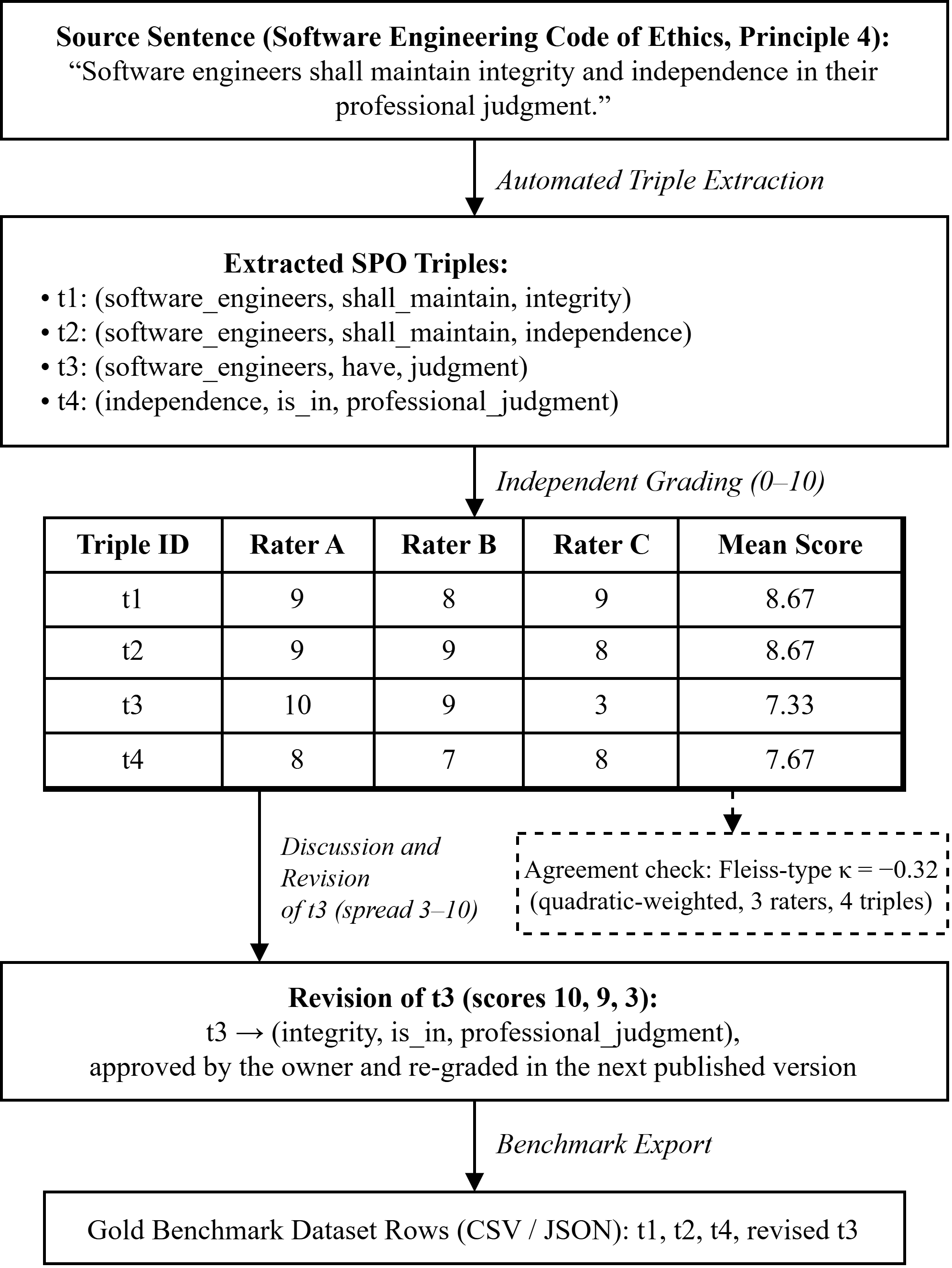}
    \caption{Worked example for the Principle 4 sentence quoted above, from extracted triples to the gold benchmark export. The triples and scores are illustrative; the kappa value is the one the platform computes for these scores.}
    \label{fig:grading_example}
\end{figure}

Fig.~\ref{fig:grading_example} follows one sentence from the ACM/IEEE-CS Software Engineering Code of Ethics and Professional Practice~\cite{gotterbarn1999code} through grading:
\begin{quote}
\emph{``Software engineers shall maintain integrity and independence in their professional judgment.''}
\end{quote}

For these scores the platform reports a quadratic-weighted Fleiss-type kappa of $-0.32$ (three raters, $n = 4$ triples). The value is low even though raters agree closely on t1, t2, and t4. Eleven of the 12 grades fall between 7 and 10, which leaves little room for chance disagreement. The single large disagreement on t3 therefore dominates the coefficient. With four items the value is unstable, and we report it only as an illustration.

\subsection{Benchmark Construction Protocol}
\label{subsec:benchmark_construction}
We recommend building a benchmark in three stages (Fig.~\ref{fig:pipeline}):

\begin{enumerate}
    \item \textbf{Stage 1: Independent grading}: graders score the candidates in the blind first round described above. This way, each grader's first grade is given without seeing other graders' scores.
    \item \textbf{Stage 2: Discussion and revision}: the owner reviews the per-triple scores and means to find triples with low scores or a wide spread. These triples are discussed in the comment threads, where graders propose revisions and vote on them before the owner approves one. Publishing the next version drops the grades and sentence-level ratings that no longer match the revised triples, as described above. The revised triples are then graded again.
    \item \textbf{Stage 3: Finalization and export}: graders request the removal of rejected triples, and each request the owner approves deletes its triple. The owner then publishes a final named version and exports it. A separate gold export (JSON or CSV, available to the owner through the REST API) holds the triples of the active version in their approved wording, along with approved additions. Triples awaiting a removal decision are left out. Membership in the gold export therefore depends on these decisions, which the grades only guide.
\end{enumerate}

Fig.~\ref{fig:grading_example} also shows how the scores lead to revision. For each triple, the platform computes the mean of the graders' scores and the spread between the lowest and the highest score. Triples t1 and t2 have a mean of 8.67 and a spread of one point. The graders agree that both are strong. Such triples are likely to be kept as they are. The mean for t4 is slightly lower, at 7.67, and it receives a short check in Stage 2. For t3, the mean is 7.33, but the grades spread from 3 to 10. Rater C gave it a 3 because the predicate \emph{have} drops the obligation ``shall maintain'' and the triple loses what must be maintained. In discussion, a grader proposes \emph{(integrity, is\_in, professional\_judgment)}, which mirrors t4. Read with t1, it states that integrity, like independence, is maintained in professional judgment, as the sentence says. Once the owner approves the proposal and publishes a new version, the revised triple is graded again.

\subsubsection{Interpreting the Scores}
The scores describe how strong or weak a triple is, and the team decides what happens to it. High scores from all graders mark a strong triple, which is likely to be kept as it is. Lower scores mark a weak triple that is more likely to be revised. When the grades sit near the bottom of the scale, removal becomes the most likely outcome. Even then, the team may revise the triple and keep it. A wide spread between graders also sends a triple to discussion.

\subsubsection{Agreement and Validity}
Kappa summarizes how consistently the graders apply the grading criteria across a session~\cite{artstein2008inter}. Because the weights are quadratic, a one-point difference between two grades counts far less than a large one (Section~\ref{subsec:irr}). A low value can point to an ambiguous sentence or an unclear guideline, which the team can clarify before the next round. Graders can also agree because they share the same misunderstanding of a guideline. Kappa therefore does not show whether a triple is true, and TripleGrader does not use it to decide membership in the benchmark. The triples to discuss are found from their own scores and spread.

\subsection{Development and Field Use}
\label{subsec:development}
During early development, we tested prototypes internally on sample sentence sets and checked persistence, file parsing, and authentication. Once the core workflow was stable, two domain graders used the tool for three months to review extracted software ethics triples. Their feedback shaped the following features:
\begin{itemize}
    \item \textbf{Session ownership}: the session list separates the owner's own sessions from sessions shared with them.
    \item \textbf{Live averages}: per-triple and session averages update as soon as a grade is entered.
    \item \textbf{Grading progress}: in the session list, a grader sees which shared sessions they have already worked on and how many graders have started grading. The workspace sidebar filters sentences into fully graded and pending.
    \item \textbf{Input checks}: uploads are checked for formatting errors before a session is created, and the parsed sentences are shown for review first.
\end{itemize}

The automated tests used in the evaluation are described in Section~\ref{subsec:eq1}.

\section{Evaluation}
\label{sec:evaluation}

Our evaluation is built around three evaluation questions (EQs). EQ1 covers workflow tests, EQ2 recovery from controlled errors, and EQ3 human curation of LLM-generated candidates into an existing benchmark. The repository holds the scripts, frozen inputs, and curation sessions needed to reproduce the reported results.

\subsection{Workflow Tests (EQ1)}
\label{subsec:eq1}

Eighteen scenarios (T01 to T18) call the backend through its HTTP interface with local file storage. Fourteen perform a permitted operation and check that it succeeds and the stored session shows the expected change. The other four attempt a prohibited operation, where the expected result is a server refusal and an unchanged session.

\begin{table*}[t]
\caption{Workflow Scenarios and Checked Behavior (EQ1)}
\label{tab:eq1_workflow}
\centering
\footnotesize
\begin{tabularx}{\textwidth}{@{}llc>{\raggedright\arraybackslash}X@{}}
\toprule
\textbf{Component} & \textbf{Scenarios} & \textbf{Tests} & \textbf{Checked behavior} \\
\midrule
Grades and sentence ratings & T02--T05 & 4 & Each grade and sentence rating is stored under its grader's ID; clearing one grade keeps the others \\
Suggestions, votes, approvals & T06--T10 & 5 & Suggestion stored with its author; self-vote and non-author deletion rejected (HTTP 403); owner approval and rollback \\
Removal and additions & T11--T12 & 2 & Triples marked for removal leave the gold export; added triples enter it only after approval \\
Comments and versions & T13--T15 & 3 & Comment and reply persist; draft edits leave the published version unchanged; only the owner can delete a version, and a session always keeps at least one \\
Creation, export, robustness & T01, T16--T18 & 4 & JSON session creation; gold JSON and CSV list the same triples; two graders' saves both kept; unknown ID rejected \\
\midrule
\textbf{Total} & & \textbf{18} & 14 permitted, 4 prohibited \\
\bottomrule
\end{tabularx}
\end{table*}

Table~\ref{tab:eq1_workflow} lists the scenarios and the behavior each one checks. Since T15 tests two, the four rejection scenarios cover five prohibited actions. These are a self-vote, a non-author deletion, and a version deletion by a non-owner (HTTP 403), deletion of the last remaining version (HTTP 400), and a suggestion for a non-existent triple (HTTP 404). In addition to the 18 scenarios, 54 tests (72 in all) cover the browser's save requests, access control, version handling, uploads, and the agreement computation. Twelve of them send the same save requests as the browser for comments, suggestions, votes, new triples, removal requests, approvals, and publishing. Eleven access-control tests check that sign-in is required and that owner-only actions are reserved for the owner. One of them also checks that peer scores stay hidden until a grader has rated an item. Twenty-two further tests cover versioning, approvals, account rules, and uploads. These include a check that a revised triple is graded again in the next version. Nine kappa tests check that the two-rater value agrees with scikit-learn's quadratic-weighted Cohen's kappa~\cite{pedregosa2011scikit} on the test cases and that the Fleiss-type value agrees with hand-computed cases. The browser's values are also checked against the server's. We did not test the contents of the Excel, Word, and PDF exports or truly concurrent writes.

\subsection{Controlled Error Injection and Recovery (EQ2)}
\label{subsec:eq2}

EQ2 tests whether the repair for each of 10 error types can be made through the REST API. As the reference, we used a benchmark of 316 triple rows (309 distinct SPO tuples) that the authors built for 38 sentences of the Code of Ethics~\cite{gotterbarn1999code}. We injected 10 error types (E1 to E10) at 10\% density with one seed. Each condition alters or removes 32 rows (E1 to E5, E8 to E10) or adds 32 spurious or duplicate rows (E6, E7). E1 to E3 replace the subject, predicate, or object, E4 swaps subject and object, and E5 deletes a row. E6 adds a spurious row and E7 a duplicate, while E8 empties the predicate and E9 respells the subject. The last type, E10, replaces subject and predicate, and its repair includes a competing wrong suggestion and a peer vote. An oracle derived the inverse actions from each injection manifest and issued them under the users' own credentials as grader suggestions, peer votes, owner approvals, removal requests, and restorations. The oracle issued 608 curation actions. \emph{Repair recall} is the share of injected changes that the exported gold set reverts, and \emph{preservation} is the share of SPO tuples untouched by the injection that the gold set keeps.

We compared the exported gold set with the reference both as a set of SPO tuples and at the row level, including polarity, modality, and note. In each condition, the gold set reverted the 32 injected changes and kept the 277 untouched tuples (309 for E6 and E7). The row-level comparison found no differences. Because the oracle knows the injected errors, EQ2 tests only whether the workflows can express each repair. Finding out whether people detect such errors would need a user study.

\subsection{Curating LLM Candidates into an Existing Benchmark (EQ3)}
\label{subsec:eq3}

The \emph{software ethics} set has 15 units from the Preamble and Principle 4 of the same Code of Ethics, and its 21 demonstrations come from Principles 7 and 8. The \emph{biomedical} set has 40 units with causal assertions from the BioCreative V Track 4 corpus of Biological Expression Language (BEL) statements~\cite{fluck2016training}, with 30 demonstrations from the same corpus. No demonstration is a test unit. We gave four models the same prompt, one unit at a time. GPT-OSS 120B and Qwen 3.8 27B were queried through the Groq API at temperature 0, and ChatGPT-5.6-High and Gemini-3.1-Pro through their web chat interfaces. Each BEL statement was split into subject, predicate, and object at the causal relation.

Both corpora already had a gold benchmark, $B$, before the candidates were extracted. For software ethics, $B$ is a 66-triple benchmark that the two authors had curated in TripleGrader. For the biomedical corpus, it consists of the 69 reference BEL statements. We uploaded each model's candidates as a separate session, eight sessions in total. The first author graded the biomedical candidates and the second author the software ethics candidates, on the 0 to 10 scale and with $B$ at hand. After the authors discussed the low-graded candidates, each candidate that did not match $B$ was corrected in the platform. A wrong slot led to a revision suggestion, a candidate with no corresponding reference triple to a removal request, and a missed reference relation to a new triple. The session owner approved each action and published the result as a new version. Since the curation targeted $B$, each published version is the same as $B$ sentence by sentence. EQ3 therefore measures the editing effort that each model's output required.

Table~\ref{tab:eq3_alignment} compares each model's frozen output with the published version of its session by candidate identifier. A candidate is \emph{exact} if it is still present with the same subject, predicate, and object, \emph{modified} if it is present but at least one slot changed, and \emph{deleted} if it is no longer present. \emph{Newly suggested} triples are those added during curation.

\begin{table*}[t]
\caption{Curation Outcomes of LLM Candidates Against the Existing Benchmark $B$ (EQ3)}
\label{tab:eq3_alignment}
\centering
\footnotesize
\begin{tabular*}{\textwidth}{@{\extracolsep{\fill}}llccccc}
\toprule
\textbf{Corpus} & \textbf{Model} & \textbf{Candidates} & \textbf{Exact} & \textbf{Modified} & \textbf{Deleted} & \textbf{Newly suggested} \\
\midrule
Software ethics & ChatGPT-5.6-High & 59 & 11 (18.6\%) & 37 (62.7\%) & 11 (18.6\%) & 18 (27.3\%) \\
($|B| = 66$) & Gemini-3.1-Pro & 49 & 10 (20.4\%) & 35 (71.4\%) & 4 (8.2\%) & 21 (31.8\%) \\
 & GPT-OSS 120B & 47 & 12 (25.5\%) & 25 (53.2\%) & 10 (21.3\%) & 29 (43.9\%) \\
 & Qwen 3.8 27B & 60 & 13 (21.7\%) & 36 (60.0\%) & 11 (18.3\%) & 17 (25.8\%) \\
\midrule
Biomedical BEL & ChatGPT-5.6-High & 60 & 19 (31.7\%) & 34 (56.7\%) & 7 (11.7\%) & 16 (23.2\%) \\
($|B| = 69$) & Gemini-3.1-Pro & 67 & 21 (31.3\%) & 37 (55.2\%) & 9 (13.4\%) & 11 (15.9\%) \\
 & GPT-OSS 120B & 68 & 7 (10.3\%) & 47 (69.1\%) & 14 (20.6\%) & 15 (21.7\%) \\
 & Qwen 3.8 27B & 65 & 7 (10.8\%) & 36 (55.4\%) & 22 (33.8\%) & 26 (37.7\%) \\
\bottomrule
\multicolumn{7}{@{}p{\textwidth}@{}}{\scriptsize Exact, modified, and deleted are shares of each model's candidates. Newly suggested is the number of triples added during curation, with its share of $|B|$. Exact $+$ modified $+$ newly suggested $= |B|$ for every model.}
\end{tabular*}
\end{table*}

Few candidates were kept unchanged, with 18.6\% to 25.5\% on the software ethics set and 10.3\% to 31.7\% on the biomedical set (Table~\ref{tab:eq3_alignment}). Most were kept but corrected, between 53.2\% and 71.4\% of each model's output. Between 4 and 22 candidates per model were removed. On the software ethics set, 35 of the 133 modified candidates had their subject, predicate, and object rewritten, against 3 of 154 on the biomedical set. Because the models also missed part of each benchmark, the grader added 17 to 29 triples per model on the software ethics set (25.8\% to 43.9\% of $B$) and 11 to 26 on the biomedical set (15.9\% to 37.7\%).
Each corpus was graded by a single author with the benchmark in view. EQ3 therefore gives no evidence on agreement between graders. Part of each modified count comes from the naming conventions of $B$, such as identifier suffixes and BEL namespaces. We also found that temperature 0 did not make the Groq outputs repeatable. A second run reproduced between 34\% and 97\% of each model's candidates. The counts in Table~\ref{tab:eq3_alignment} each describe a single run.

\section{Discussion}
\label{sec:discussion}

TripleGrader keeps the grades, comments, and revisions of a review with its session. The data model is a generic subject-predicate-object triple. For that reason, the same workflow could support expert review of biomedical relations or of obligations in codes and standards. It could also provide the human grading step of benchmarks such as Text2KGBench~\cite{mihindukulasooriya2023text2kgbench}. A team can start sessions from CSV, Excel, or JSON files without writing code, while extraction pipelines can create sessions and collect gold exports through the REST API. The scale or the weight matrix can be changed in the open-source code.

For all four models in EQ3, the modified candidates were the largest group. Most candidates were kept after a correction, usually to one or two slots. An accept or reject label would hide this partial correctness, which an ordinal grade records. With revision and re-grading, the team can fix the wrong slot and keep the triple.

So far, we have not run a controlled user study comparing annotation speed or workload with other workflows. The agreement measures also still await data from several independent graders. EQ3 measures only how much editing these models' output needed on these corpora.

\section{Conclusion}
\label{sec:conclusion}

We presented TripleGrader, a web platform for grading pre-extracted triples and building benchmark datasets through multi-annotator review. It covers the grading cycle from upload to export. During independent grading, peer scores are withheld. When the owner approves a revision, it enters a new version in which the revised triple is graded again. The platform also computes quadratic-weighted kappa and produces a separate gold export.

We evaluated selected grading and revision workflows using automated API-level tests and controlled error-injection checks. The repair tests assessed whether prescribed corrections could be applied while preserving unaffected records. On two small corpora, most candidates from four LLMs needed correction before they matched the existing gold benchmarks, partly because of those benchmarks' naming conventions. Triples extracted by these models therefore still needed human grading and review before they could serve as a benchmark. A user study and grading by several independent graders are still missing, and we plan both as next steps. The source code is available at \url{https://github.com/razz0n/triple-grader-tool}.

\section*{Acknowledgment}
The authors thank the BioCreative V BEL task organizers and corpus contributors for making the reference corpus available for research. ChatGPT (OpenAI) and Claude (Anthropic) assisted with tool development and with generating evaluation scripts according to the authors' design. Both tools also helped draft and revise every section of this paper. The authors reviewed and edited the AI-assisted material and take full responsibility for the study design, data curation, analyses, and conclusions.

\bibliographystyle{IEEEtran}
\bibliography{references}

\end{document}